\documentclass[letterpaper,10pt,conference]{ieeeconf}

\IEEEoverridecommandlockouts
\usepackage{amsmath}
\usepackage{amssymb}
\usepackage{booktabs}
\usepackage{balance}
\usepackage{float}
\usepackage{graphicx}
\usepackage{makecell}
\usepackage{tabularx}
\usepackage{tikz}
\usepackage{url}
\usepackage[hidelinks]{hyperref}
\usetikzlibrary{arrows.meta,positioning,fit}

\newcommand{\original}{\textsc{Original}}
\newcommand{\taskctx}{\textsc{TaskCtx}}
\newcommand{\stagestate}{\textsc{Ordinal Stage-State}}

\title{An Empirical Study on Stage-Information Interfaces\\
for VLA Fine-Tuning}

\author{Yingwei Ji}

\begin{document}
\maketitle
\thispagestyle{empty}
\pagestyle{empty}

\begin{abstract}
One high-level instruction in long-horizon manipulation can cover several action stages.
We use segmented action annotations as an intermediate representation between the full-task instruction and VLA action chunks.
A progress module tracks the active stage, while the action policy receives stage information either as current-stage text or as a normalized ordinal stage index in robot state.
We compare these interfaces with GR00T N1.6 on LIBERO-10 under direct fine-tuning and continuation fine-tuning from a full-task instruction baseline.
Under direct fine-tuning, full-task instruction, current-stage text, and \stagestate{} achieve mean success rates of 57.45\%, 50.24\%, and 54.36\%, respectively, showing that explicit stage information does not automatically improve the policy.
Under continuation, the corresponding means are 49.07\%, 50.00\%, and 53.75\%, with \stagestate{} exceeding both alternatives in all three paired runs.
The observed benefit differs across interface representations and training arrangements.
\end{abstract}

\section{Introduction}

Long-horizon manipulation maps one stable task goal to a changing sequence of low-level actions.
For example, the task \emph{put both moka pots on the stove} includes approaching the first pot, grasping it, transporting it, placing it, and repeating the sequence for the second pot.
We began with a practical intuition: if the task is divided into subtasks, the language can describe the current action more precisely, and the action distribution within each segment should become narrower.
With less variation in each segment, a vision-language-action (VLA) policy might be easier to train.

Our results did not follow this intuition uniformly.
In direct fine-tuning, both the stage-text and ordinal-state variants were below the \original{} mean.
Adding a segment signal did not automatically improve the policy, and in some cases it reduced rollout success.
That mismatch brings us back to a more basic question than whether segmentation is useful: how should subtask information be represented and introduced during action-policy training?

One possible explanation is that stage conditioning changes how examples share representations across phases.
The \original{} task instruction keeps one language representation for all stages, while observations and robot state distinguish the current behavior.
\taskctx{} instead appends the active stage phrase to the task prompt, supplying a different language condition for each stage.
Most vision-language parameters are frozen, while the top four language-model layers remain trainable.
\taskctx{} changes the language condition seen by both these upper layers and the action module using limited task data.

We study the interface question on LIBERO-10 with GR00T N1.6.
A progress module estimates completion of the active stage and controls when the system advances.
The resulting stage pointer is exposed to the action policy either as active-stage text or as an ordinal stage index in state.
We compare these designs under two arrangements: direct fine-tuning from the pretrained model and continuation from a full-task baseline.
The comparisons use the same stage annotations, controller, GR00T backbone, fine-tuning scope, and paired action-policy runs.

We make two contributions.
First, we formulate subtask-conditioned VLA fine-tuning as an interface problem between a stage-tracking module and an action policy, comparing current-stage text with an ordinal stage-state signal under shared annotations and control logic.
Second, we evaluate the two interfaces under direct fine-tuning and continuation from a full-task baseline.
Explicit stage information does not exceed the \original{} mean under direct training, while \stagestate{} achieves the highest continuation mean and exceeds both alternatives in all three paired runs.

\begin{figure*}[t]
\centering
\includegraphics[width=0.94\textwidth]{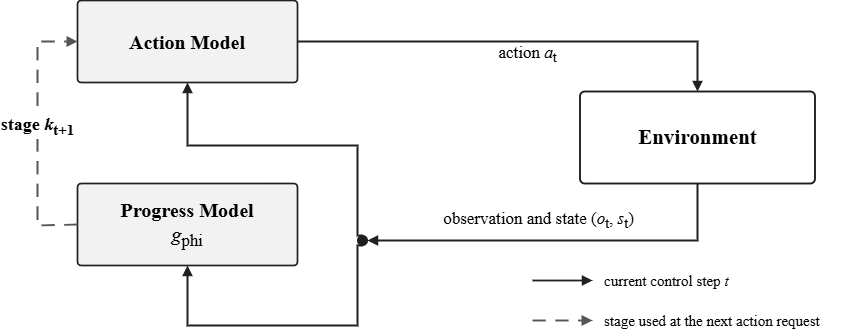}
\caption{Closed-loop rollout pipeline. A single action model produces the current action chunk. While the environment executes that chunk, the progress model updates the stage controller for the next request. The action model is treated as one module; only its stage-information input differs across experiments.}
\label{fig:pipeline}
\end{figure*}

\section{Related Work}

Recent VLA systems have made imitation-based robot learning increasingly capable.
RT-2 connects web-scale vision-language knowledge to robotic control, while OpenVLA, Octo, RoboFlamingo, and SmolVLA explore open or efficient adaptation across manipulation settings~\cite{rt2,openvla,octo,roboflamingo,smolvla}.
Continuous-action systems such as $\pi_0$ and GR00T N1 further support direct action generation with flow- or diffusion-style heads~\cite{pi0,groot}.
However, stronger models do not remove a practical bottleneck: collecting diverse robot demonstrations remains expensive.
Adding descriptions to trajectories that have already been collected can be substantially cheaper than gathering new physical interactions, and modern vision-language models can help propose detailed action segments for review.
This creates an appealing possibility: extract more supervision from fixed robot data by describing what the robot is doing at each stage.

Long-horizon systems already suggest that intermediate language can organize behavior.
SayCan connects high-level reasoning to executable skills, RT-H inserts language motions between task instructions and low-level actions, and DeMiAn enriches demonstrations with segment-level descriptions~\cite{saycan,rth,demian}.
These works make detailed subtask annotation technically plausible, but they do not make its benefit automatic.
Once a trajectory has been divided into stages, the action policy still needs an interface through which that information can affect learning.
Our question comes from the practical use of annotation rather than annotation generation itself: does an active-stage label help a fine-tuned VLA, and should it enter through language or state?

Progress-aware systems provide a natural way to keep track of the active stage.
ProgVLA and ProgressVLA estimate task progress, SeqVLA uses completion to trigger transitions, and CycleVLA and See-Plan-Rewind use progress or milestones for monitoring and recovery~\cite{progvla,progressvla,seqvla,cyclevla,sprvla}.
We use completion in this internal role: a progress model advances a shared stage pointer, but its completion scalar is not given to the action policy.
The experiment focuses on a concrete module-interface choice under the same stage tracker and annotations: expose the active stage as semantic text, or as a normalized ordinal value in robot state.
We study this choice by fine-tuning GR00T N1.6 with its visual encoder and most language-model layers frozen~\cite{grootn16}.

\section{Method}

Let $o_t$ be the visual observation, $s_t$ the robot state, and $l_{\mathrm{task}}$ the original task instruction.
The action module predicts a low-level chunk $a_{t:t+H}$.
We insert an ordered action-stage representation between $l_{\mathrm{task}}$ and this chunk.
Figure~\ref{fig:pipeline} shows how the progress and action models form a closed loop with the environment, with the estimated stage used by the next action request.

\begin{figure*}[t]
\centering
\includegraphics[width=0.88\textwidth]{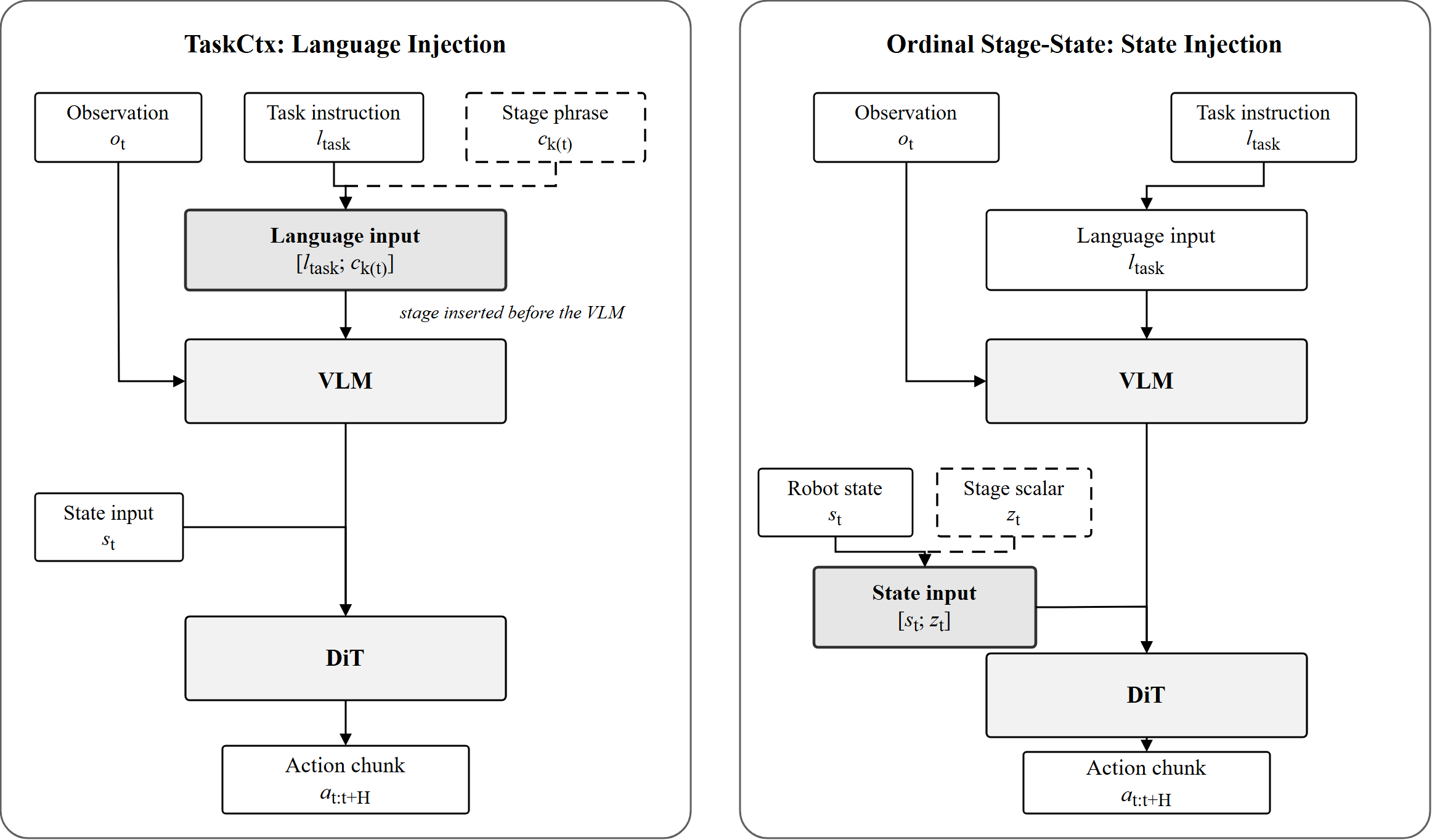}
\caption{Stage-information injection routes. Each experiment selects one mapping from the shared stage pointer to the inputs of the same action-model module. \taskctx{} changes language; \stagestate{} appends the normalized ordinal value to robot state.}
\label{fig:injection}
\end{figure*}

\subsection{VLA Backbone and Fine-Tuning Scope}

We use the pretrained NVIDIA GR00T N1.6-3B model~\cite{groot,grootn16}.
Its VLM encodes camera images and task language.
State and action projectors connect robot inputs to a 32-layer Diffusion Transformer (DiT), which predicts relative action chunks with flow matching.
We freeze the visual encoder and all but the top four language-model layers.
We fine-tune those four language layers together with the action-side projectors, vision-language layer normalization, and DiT on LIBERO.

\subsection{Action Segmentation}

For each LIBERO task, Qwen2.5-VL~\cite{qwen25vl} receives the original task information and instructions for action segmentation.
It produces the ordered stage phrases and demonstration frame ranges used in this study.
We use one fixed set of stage phrases and demonstration frame boundaries for all methods.
The annotation procedure is not an experimental variable; we compare how the resulting stage information is exposed to the action module.
These frame ranges label demonstrations and are not provided as an oracle during rollout.
Adjacent low-level motions are grouped into semantic stages, so one stage can include approach, grasp, transport, and placement motions when they serve the same subgoal.
The resulting sequence $C=(c_1,\ldots,c_K)$ lies between the complete task instruction and low-level action chunks.

\begin{table}[H]
\caption{Example stages from one annotated demonstration.}
\label{tab:stages}
\centering
\scriptsize
\setlength{\tabcolsep}{3pt}
\begin{tabularx}{\columnwidth}{@{}lXrl@{}}
\toprule
Field & Example & Index & Time / frames \\
\midrule
Task & Turn on the stove and put the moka pot on it & -- & 27.2 s / 0--272 \\
Stage & PressButton the stove & 1 & 0--10.5 / 0--105 \\
Stage & Place the moka pot on the stove & 2 & 10.5--25 / 105--250 \\
Stage & Retract arms & 3 & 25--27.2 / 250--272 \\
\bottomrule
\end{tabularx}
\end{table}

Table~\ref{tab:stages} shows three stages from the fixed annotation used throughout the study.
Frame boundaries vary across demonstrations.

\subsection{Progress Module}

Frame boundaries $[b_k,e_k]$ assign each training frame to stage $k$.
The local target is
\begin{equation}
p_t=\operatorname{clip}\left(\frac{t-b_k}{e_k-b_k},0,1\right).
\end{equation}
The progress model $g_\phi$ is initialized from the pretrained GR00T checkpoint and receives $o_t$, $s_t$, $l_{\mathrm{task}}$, and the active phrase $c_k$.
During progress training, the visual encoder and all but the top four language-model layers are frozen.
The top four language layers, action-side projector, diffusion/DiT module, and action-head VLLN remain trainable, while a linear scalar head followed by a sigmoid predicts progress.
We optimize Smooth L1 loss, disable the action loss, and use zero action tokens during progress-only training.

At rollout, the system maintains a stage pointer $k_t$ and queries
\begin{equation}
\hat p_t=g_\phi(o_t,s_t,[l_{\mathrm{task}};c_{k_t}]).
\end{equation}
The pointer advances when $\hat p_t>0.8$ after at least 15 stage frames.
The pointer is exposed to the action policy either as the active stage label or as its normalized ordinal index.
The completion estimate only advances the pointer and is not provided to the action policy.

\subsection{Action-Module Interfaces}

Figure~\ref{fig:injection} summarizes the two alternative mappings from the shared stage pointer to the action-model inputs.

\textbf{Original.}
The comparison receives no explicit stage input:
\begin{equation}
a_{t:t+H}=\pi_A(o_t,s_t,l_{\mathrm{task}}).
\end{equation}

\textbf{Semantic Stage Interface (\taskctx{}).}
The active phrase is sent through language:
\begin{equation}
a_{t:t+H}=\pi_B(o_t,s_t,[l_{\mathrm{task}};c_{k_t}]).
\end{equation}

\textbf{Ordinal Stage-State Interface.}
The original instruction is preserved, and the active-stage index is appended to state:
\begin{equation}
a_{t:t+H}=\pi_C(o_t,[s_t;z_t],l_{\mathrm{task}}).
\end{equation}
The progress module controls the one-based pointer $k_t\in\{1,\ldots,K_\tau\}$ for task $\tau$, which the processor maps to $z_t$.
Let $K_{\max}=\max_\tau K_\tau$; for LIBERO-10, $K_{\max}=6$.
GR00T applies min-max state normalization:
\begin{equation}
z_t=2\frac{k_t-1}{K_{\max}-1}-1.
\end{equation}
Thus, the action model receives the clipped value $z_t\in[-1,1]$, not $k_t$ and not a task-relative value normalized by $K_\tau$.
Appending $z_t$ expands the active dataset and processor state schema from eight to nine dimensions.
The GR00T state encoder already contains the corresponding ninth input column, so no model tensor is resized.
All projector weights, including the nonzero ninth column, are restored from the same source checkpoint within each comparison.
The original eight-dimensional LIBERO interface does not activate the ninth slot during baseline fine-tuning.

\begin{table}[t]
\caption{Inputs received by the action module. Here $g_L,g_R$ are the two finger-state dimensions.}
\label{tab:interfaces}
\centering
\scriptsize
\setlength{\tabcolsep}{3pt}
\begin{tabularx}{\columnwidth}{@{}lXX@{}}
\toprule
Method & Text & State \\
\midrule
\original{} & task & $x,y,z,\mathrm{rpy},g_L,g_R$ \\
\taskctx{} & task + stage & $x,y,z,\mathrm{rpy},g_L,g_R$ \\
\stagestate{} & task & $x,y,z,\mathrm{rpy},g_L,g_R,z_t$ \\
\bottomrule
\end{tabularx}
\end{table}

\subsection{Training Arrangements}

We evaluate two training arrangements.
Direct fine-tuning introduces each interface while adapting the pretrained GR00T policy to the downstream tasks.
Baseline-first continuation first adapts the policy with the complete task instruction and then introduces each interface from the resulting full-task checkpoint.
Within each arrangement and paired run, \original{}, \taskctx{}, and \stagestate{} share the same source initialization and training budget.

\subsection{Closed-Loop Deployment}

The progress and action models run as a pipeline (Fig.~\ref{fig:pipeline}).
While the environment executes the current 8-step action chunk, the progress module estimates active-stage completion.
Its output updates the stage pointer and conditions the next action-chunk request.
Progress inference runs in parallel and completes before the next request.

\section{Experiments}

\subsection{Experiment Setup}

\textbf{Benchmark and data.}
We evaluate on LIBERO-10~\cite{libero}.
One action model is trained on all ten tasks using the first 30 demonstrations per task.
All experiments use the LIBERO/GR00T eight-dimensional state and seven-dimensional action formats described above.
\stagestate{} appends one normalized scalar to state.

\textbf{Models.}
We train the original-language baseline and two stage-conditioned direct variants from pretrained GR00T and evaluate them at 5k steps.
For each training seed, continuation variants start from the same corresponding full-task baseline-5k checkpoint and continue for 1k steps.
A single fixed progress model is trained for 10k steps and shared across all action-policy seeds and both arrangements.
Its completion estimate controls the stage pointer but is not sent to the action policy.

\textbf{Compute and rollout.}
Training and rollout use a server with four NVIDIA A800 GPUs.
Each method is evaluated on all 10 tasks with 50 requested episodes per task and $n_{\mathrm{envs}}=10$.
We use rollout seed 7001, action and progress strides of 8, threshold 0.8, and at least 15 stage frames before advancement.
The actual episode count is read from one CSV row per completed episode.
Each arrangement is repeated in three paired action-policy runs with fixed random seeds 101, 202, and 303.

\subsection{Direct Fine-Tuning}

\begin{table}[t]
\caption{Direct 5k success rate. Mean $\pm$ sample standard deviation is across three paired action-policy runs.}
\label{tab:direct}
\centering
\scriptsize
\setlength{\tabcolsep}{3.2pt}
\begin{tabular}{@{}lrrrr@{}}
\toprule
Method & Run 1 & Run 2 & Run 3 & Mean $\pm$ std \\
\midrule
\original{} & 64.80 & 58.96 & 48.60 & \textbf{57.45 $\pm$ 8.20} \\
\taskctx{} & 44.62 & 49.00 & 57.09 & 50.24 $\pm$ 6.32 \\
\stagestate{} & 45.80 & 62.08 & 55.20 & 54.36 $\pm$ 8.17 \\
\bottomrule
\end{tabular}
\end{table}

Table~\ref{tab:direct} answers whether explicit stage information helps when introduced at the start of downstream adaptation.
Neither stage-conditioned interface exceeds the \original{} mean.
\stagestate{} is 4.12 percentage points above \taskctx{} and exceeds it in two of the three paired runs, but remains 3.09 points below \original{} in mean.
\taskctx{} is 7.21 points below \original{} in mean, although it is higher in Run 3.

\subsection{Baseline-First Continuation}

\begin{table}[t]
\caption{Success after 5k full-task baseline fine-tuning and 1k continuation. Mean $\pm$ sample standard deviation is across paired runs.}
\label{tab:continuation}
\centering
\scriptsize
\setlength{\tabcolsep}{3.2pt}
\begin{tabular}{@{}lrrrr@{}}
\toprule
Method & Run 1 & Run 2 & Run 3 & Mean $\pm$ std \\
\midrule
\original{} & 47.60 & 52.00 & 47.60 & 49.07 $\pm$ 2.54 \\
\taskctx{} & 53.89 & 49.20 & 46.91 & 50.00 $\pm$ 3.56 \\
\stagestate{} & 57.57 & 54.69 & 49.00 & \textbf{53.75 $\pm$ 4.36} \\
\bottomrule
\end{tabular}
\end{table}

Table~\ref{tab:continuation} asks which interface has the highest mean after full-task baseline fine-tuning.
The ordering is \stagestate{} $>$ \taskctx{} $>$ \original{}.
\stagestate{} exceeds \taskctx{} and \original{} on all three paired runs, improving their means by 3.75 and 4.69 percentage points, respectively.
\taskctx{} is 0.93 points above \original{} in mean and exceeds it in one run.

\subsection{Result Scope}

The reported standard deviations are across three independent paired action-policy runs, with the rollout seed fixed.
A single 10k-step progress model is shared across all action-policy runs and both arrangements.
The two absolute success-rate ranges are not combined because the arrangements use different initialization histories and training trajectories.
Each arrangement is compared internally.
\stagestate{} has a higher mean than \taskctx{} in both arrangements, but neither stage-conditioned interface exceeds \original{} under direct training.

\section{Discussion}

The direct results complicate our initial intuition that a more precise stage description should make action learning easier.
Knowing the active stage can reduce action ambiguity, but neither stage-conditioned interface exceeds \original{} under direct fine-tuning.
The interface itself becomes part of the learning problem.

\subsection{Interface Representation and Data Sharing}

The higher mean of \stagestate{} relative to \taskctx{} can be read through the existing GR00T computation path.
In the action DiT, the robot state token and noised action tokens form the hidden sequence that produces cross-attention queries, while visual and language tokens provide keys and values.

\taskctx{} modifies the prompt before vision-language encoding.
Each stage produces a different language context, and the action queries must retrieve stage semantics through cross-attention.
Because the visual encoder and most language layers are frozen, adaptation to these prompts mainly occurs in the top four language layers and the action module.

\stagestate{} keeps the full-task prompt unchanged.
The normalized stage index passes through the state projector and becomes part of the state-action sequence before the first DiT block.
It participates in self-attention from the beginning of action computation and changes the queries used in later cross-attention layers.

These paths also suggest a difference in how examples share representations.
\taskctx{} partitions one task's demonstrations across several stage-conditioned language contexts, whereas \stagestate{} retains a shared full-task language context and separates stages with a low-dimensional action-side variable.
More specific conditioning may reduce ambiguity within a stage, but it also leaves fewer examples under each condition.
This variance trade-off offers one account of why \stagestate{} has a higher mean than \taskctx{} in both arrangements while both remain below \original{} under direct fine-tuning.

\subsection{Introducing Stage Information after Full-Task Fine-Tuning}

The continuation results add a second observation: \stagestate{} exceeds both alternatives in all three paired runs after the shared full-task baseline.
During direct fine-tuning, the policy must adapt to LIBERO and learn the new stage interface at the same time.
Baseline-first continuation instead introduces stage information after the policy has learned a basic mapping from task language, observations, and robot state to actions.

For \stagestate{}, the full-task prompt learned during baseline fine-tuning remains unchanged.
The continuation phase introduces only the ordinal state variable, which may act as a compact adjustment to the existing action representation.
\taskctx{} augments the baseline prompt with stage-dependent context, requiring the upper language layers and action module to adapt the existing language-action relationship to several new contexts.
This difference in compatibility with the baseline policy offers one explanation for the continuation ordering.
It also shifts the practical question from whether stage annotations are available to how the stage-tracking module communicates with the action policy and when that interface is introduced.

\subsection{Limitations}

The study is limited to LIBERO-10, GR00T N1.6-3B, the current parameter-freezing setup, and the evaluated training budgets.
Direct and continuation use different initialization histories and update budgets.
\taskctx{} and \stagestate{} provide different forms of stage information, so the comparison does not isolate the input channel alone.
The stage annotations, progress model, and rollout seed are fixed across action-policy runs.
Three paired action-policy runs provide only a limited view of variation across training seeds.

\section{Conclusion}

We study how active subtask information should be exposed to a VLA action policy.
Under direct fine-tuning, neither stage text nor ordinal stage state exceeds the \original{} mean, showing that explicit stage annotations do not automatically simplify policy learning.
After continuation from a full-task baseline, \stagestate{} achieves the highest mean and exceeds both \original{} and \taskctx{} in all three paired runs.
These results suggest that a low-dimensional stage-state interface can be more effective than modifying the language input processed by a mostly frozen vision-language backbone, particularly when adapting an existing full-task policy.
Whether this pattern carries to matched training schedules, other benchmarks, and real robots remains open.

\balance
\bibliographystyle{IEEEtran}
\bibliography{refs}

\end{document}